\documentclass[letterpaper, 10 pt, conference]{ieeeconf}
\IEEEoverridecommandlockouts 
\usepackage{graphicx}           
\usepackage{url}
\usepackage{bm}
\usepackage{verbatim}
\usepackage{subcaption}
\usepackage{float}
\usepackage{caption}
\usepackage{todonotes}
\usepackage{soul}
\usepackage{blindtext}
\usepackage{amsmath}
\usepackage{amssymb}
\usepackage{times}
\usepackage{lipsum}
\usepackage{epstopdf}
\usepackage{mathptmx}
\usepackage[super]{nth}

\newcommand{\minus}{\scalebox{0.75}[1.0]{$-$}}
\title{\LARGE \bf
A Video-Based Activity Classification of Human Pickers in Agriculture}
\author{Abhishesh Pal$^{1}$, Antonio C. Leite$^{1}$, Jon G. O. Gjevestad$^{1}$, Pål J. From$^{1,2}$
\thanks{$^{1}$Abhishesh Pal, Antonio C. Leite, Jon G. O. Gjevestad and Pål J. From
are with the Faculty of Science and Technology, Norwegian University of       Life
Sciences, Universitetstunet 3, 1433 Ås, Norway. 
{\tt E-mail: [abhishesh.pal, pal.johan.from]@nmbu.no}}%
\thanks{$^{2}$Pål J. From
is with the Lincoln Institute for Agri-Food Technology, University of Lincoln, Lincoln LN6 7TS,
United Kingdom. 
{\tt E-mail: pfrom@lincoln.ac.uk}}%
}
\begin{document}
%
\maketitle
\thispagestyle{empty}
\pagestyle{empty}
%
\begin{abstract}
In farming systems, harvesting operations are tedious, time-                 and
resource-consuming tasks.
Based on this, deploying a fleet of autonomous robots to work          alongside
farmworkers may provide vast productivity and logistics benefits. 
Then, an intelligent robotic system should monitor human behavior, identify  the
ongoing activities and anticipate the worker's needs.
%
%
%
%
In this work, the main contribution consists of creating a benchmark       model
for video-based human pickers detection, classifying their activities to   serve
in harvesting operations for different agricultural scenarios.
%
%
Our solution uses the combination of a Mask Region-based Convolutional    Neural
Network (Mask R-CNN) for object detection and optical flow for motion estimation
with newly added statistical attributes of flow motion descriptors, named     as
Correlation  Sensitivity (CS).
A classification criterion is defined based on the Kernel Density     Estimation
(KDE) analysis and K-means clustering algorithm, which are implemented 
%
%
upon in-house collected dataset from         different
crop fields like strawberry polytunnels and apple tree orchards.
%
%
The proposed framework is quantitatively analyzed using sensitivity, specificity,
and accuracy measures and shows satisfactory results amidst various       dataset
challenges such as lighting variation, blur, and occlusions. 
%
%
\end{abstract}
%
\section{Introduction}
\begin{figure}[htpb]
    \centering
    \includegraphics[width=8.4cm, height=2.2cm]{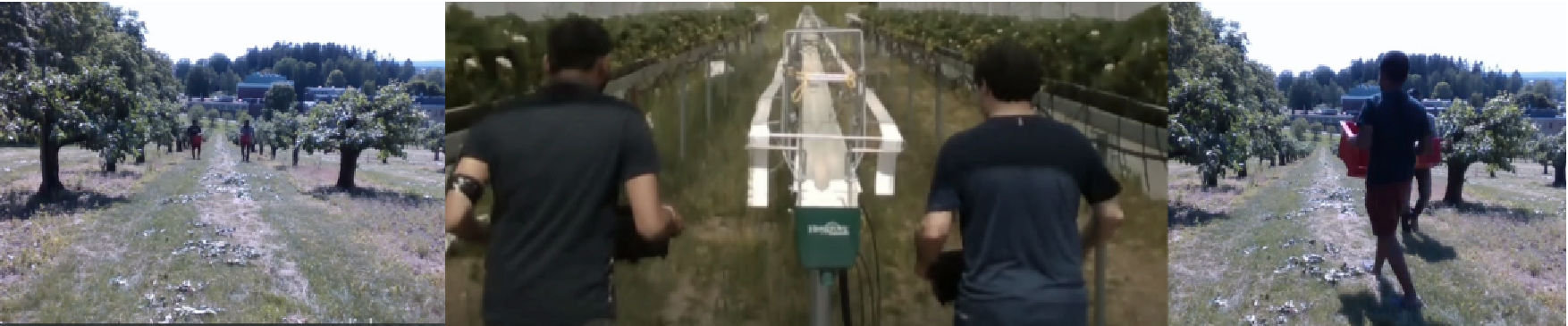}
    \caption{Dataset collecting challenges in agricultural fields: blur (left), 
      lighting  variation (center), and occlusion (right).}
    \label{fig:dataset_challenges_new}
\end{figure}

%
The agricultural field has emerged as a promising application for human activity
recognition (HAR) towards the development and optimization of efficient    robot
fleet operations and human-robot collaboration in farming and         harvesting
operations \cite{from2018rasberry}.
%
%
According to \cite{wang2019deep}, an HAR system must deal with many    technical
challenges related to the extraction of distinguishable features, scarcity    of
annotations, the complexity of data association, the heterogeneity of        the
sensory data, and the interpretability in sensory data.
%
%
The use of HAR systems in agriculture is relatively new, especially      when it
comes to recognizing human pickers' field operational activities, such        as
distinguishing between harvesting, unloading, and transportation of       trays,
where not much research is done in the past.
%
%
%
For example, training and testing of machine learning-based methods require  the
availability of large annotated data sets as well as a baseline model to compare
them. However, collecting and manually annotating sensory activity data      are
expensive and time-consuming tasks, especially in agricultural fields. 
%
%
%
%
%
The video or images captured from a real-time soft-fruit picking       operation
basically contains features of the actions and events (i.e., the duration     of
an activity). 
%
%
%
Usually, these visual features might carry different noise                sources
(Fig.\,\ref{fig:dataset_challenges_new}), such as dynamic background interference,
partial or total occlusion, and distinct viewpoints. 
Moreover, it is still a laborious task to identify motions and objects from    a
content-based video analysis. 
Then, it may be quite challenging to obtain efficient and accurate        motion
representation for humans working in agricultural fields.  
%
%

In general, an HAR process includes two main steps: feature extraction based  on
image/video information, and image/video classification based on the   extracted
feature vectors \cite{suneth2016hra}.
%
%
In static images, visual (spatial) features are for example pixel        values,
corners, blobs and edge histograms. 
Thus, a feature extraction method must compute a reliable descriptor from    the
pixels around each interest point, such as SIFT, SURF and BRIEF. 
%
%
In dynamic scenes, along with static images, visual features    also     include
temporal features as interest points, displacement and trajectories.
A spatio-temporal interest point encodes the video information at a     specific
location in space and time. 
%
%
On the other hand, a trajectory tracks a certain spatial point over time    and,
thus, captures the motion information \cite{wang2013}.

Global features describe an image as a whole entity and can generalize an entire
object within a scene with a single vector. 
In contrast, local features represent image patches and are computed at multiple
points in the image, being  more robust to occlusion and cluttering.
Thus, local space-time features may provide a successful representation for  the
recognition of human activity \cite{vasudevan2013}.
%
%
There are several feature extraction approaches for extracting             local
spatio-temporal features in videos, such as Gabor filters,       Hessian matrix,
higher-order derivatives, gradient information, optical                     flow
\cite{fleet2006optical}, and brightness information.
%
%
%
%

\begin{figure}[!htpb]
\centering
\includegraphics[height=3.5cm, width=8.4cm]{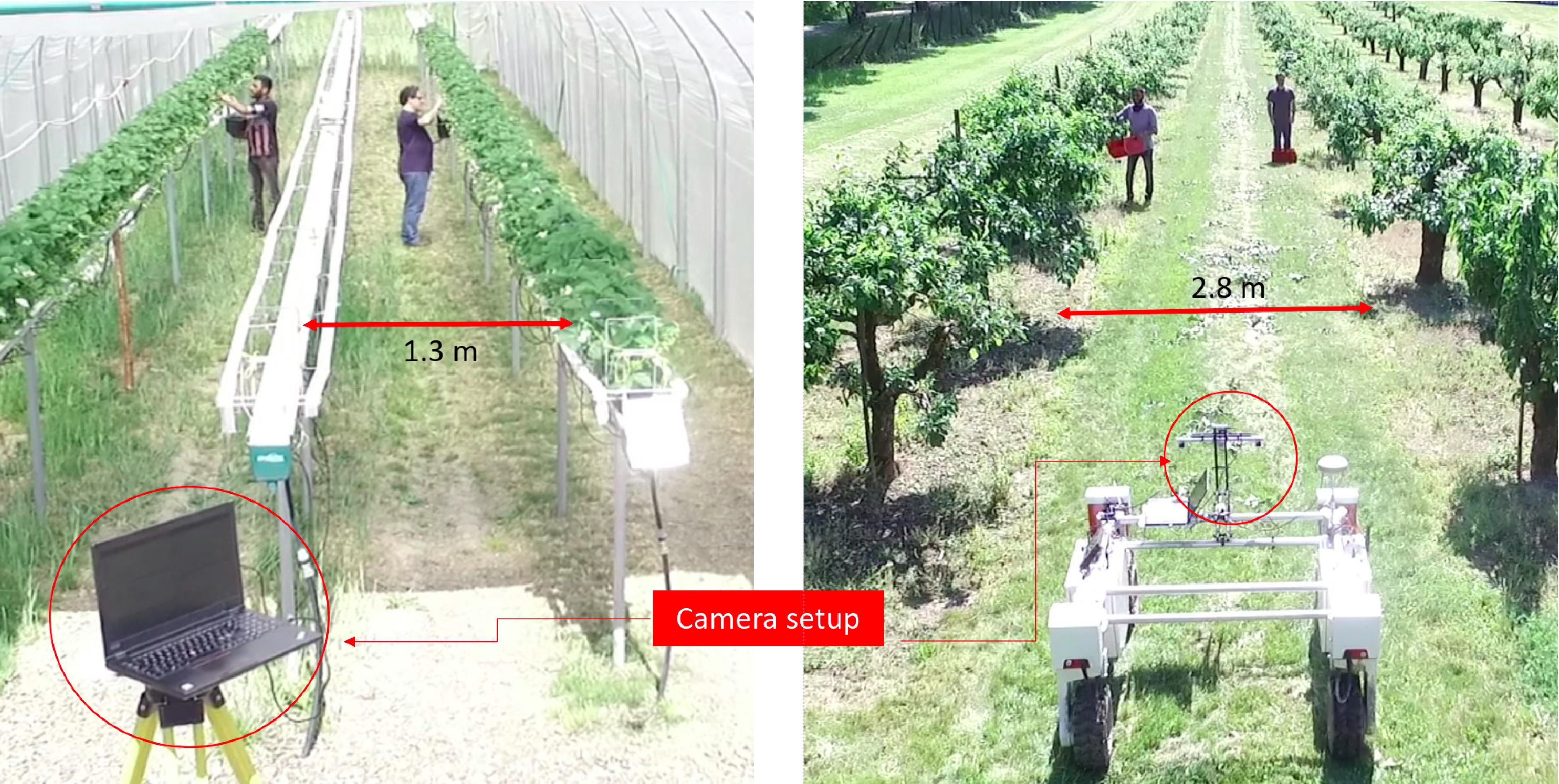}
\caption{Data collection setup in a strawberry polytunnel (left side) and    an
apple tree orchard (right side).}
\label{fig:data_collection_polyorch}
\end{figure}

\begin{figure}[htbp]
\centering
\vspace{-3mm}
\includegraphics[width=8.4cm]{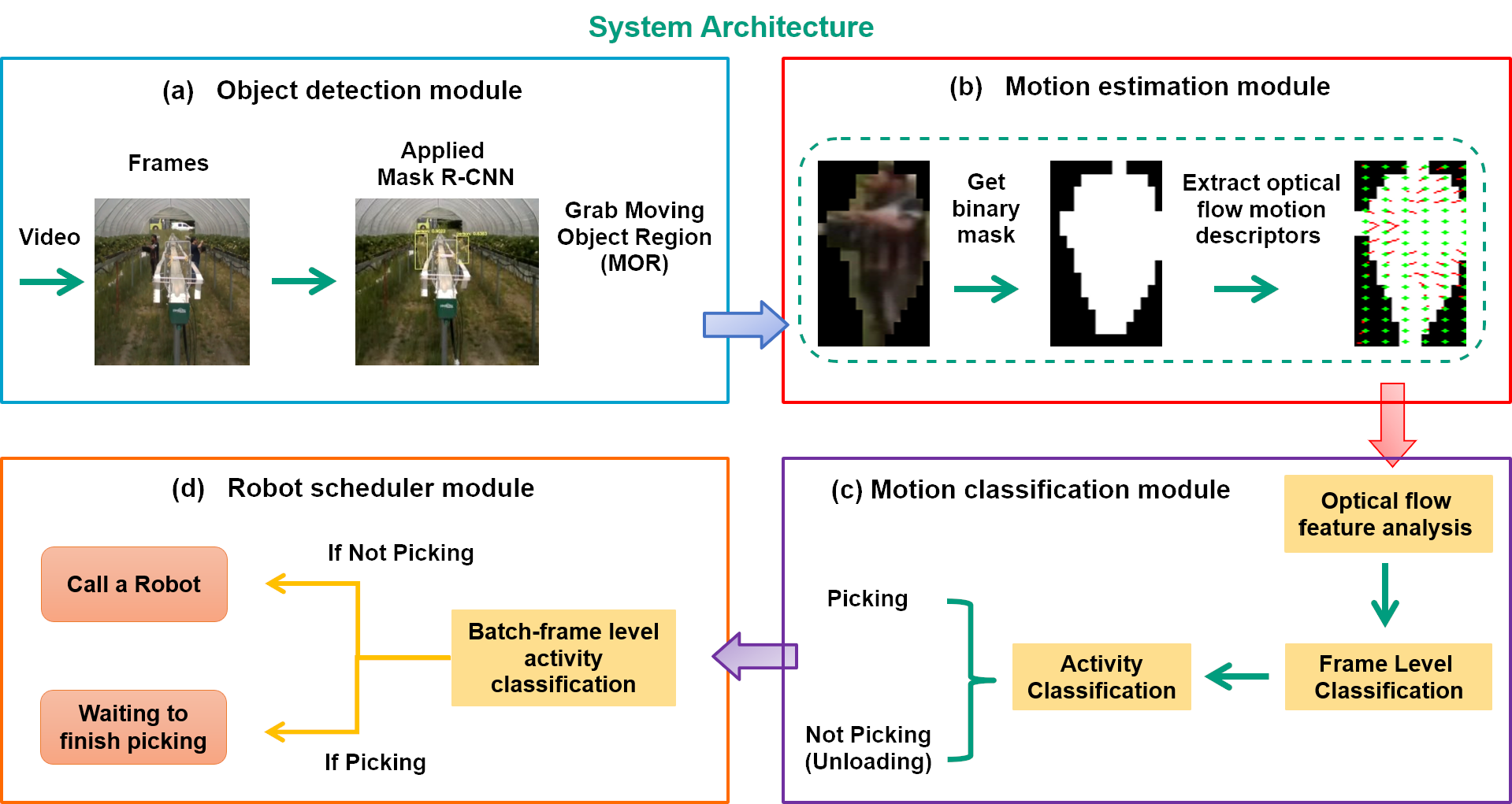}
%
\caption{Overview of the proposed video-based activity recognition and classification architecture.}
\label{fig:proposed_architecture_1}
\end{figure}

This work proposes a video-based benchmark model that identifies human   pickers
in a scene with visual noises and classifies which agricultural activities   the
identified workers are performing on the crop field. 
For this, we will extract the visual feature map from  image sequences   of  the
pickers' activities in different fruit picking scenarios 
(Fig.\,\ref{fig:data_collection_polyorch}).
%
%
The classification model consists of $4$ (four) main modules with       distinct
functions per each, namely: Object Detection Module (ODM), Motion     Estimation
Module (MEM), Motion Classification Module (MCM), and  Robot Scheduler    Module
(RSM), as shown in Fig.\,\ref{fig:proposed_architecture_1}.
%
%
In a nutshell, our classification framework uses a  mask salient          object
detection approach based on the combination of a Mask Region-based Convolutional
Neural  Network (Mask R-CNN) architecture to extract visual features and, optical
flow motion descriptors to understand behavioral activity patterns, with     the
help of newly added statistical attributes defined as \textit{mean}           of
Correlation Sensitivity  (CS) along with other attributes such as        minimum
(\textit{min.}), maximum (\textit{max.}) and \textit{range} of descriptors.
%
%

To decipher the underlying patterns in motion descriptor's           statistical
attributes and to create class labels, we explore the synergy between the Kernel
Density Estimation (KDE) analysis and the K-means clustering algorithm. 
%
%
%
We apply both methods on the same dataset, collected in distinct crop     fields,
and discuss their performance based on data characteristics. 
%
%
%
In view of that, we set the classification criteria by choosing the thresholding
parameters to the identified statistical attributes of the motion descriptor.
%
%
To evaluate the relevance of the CS attribute in pattern recognition         and
classification of activities, we compare the practical results obtained with the
proposed video-based activity classification algorithm, which           includes
\textit{mean}\, CS, against other variants of the same algorithm that does   not
have CS.
%
%
%
%
\vspace{-4.5mm}
\section{Data Collection and Challenges} \label{sec:data_collection}
%
%
%
%
The data is collected in-house with two human pickers working in different fruit
picking scenarios, viz. polytunnel and open field, located at the      Norwegian
University of Life Sciences (NMBU), Campus Ås, Norway.
The setup for data collection is shown in Fig.\,\ref{fig:data_collection_polyorch},
where we have installed the                        Intel{\tiny{\textregistered}}
RealSense{\texttrademark} depth camera D435   (pixel size: $3\mu\!\times\!3\mu$;
focal length: $1.93\,mm$; RGB frame rate: $30\,\mbox{fps}$) at an    approximate
height of $1\minus1.5\,\mbox{m}$ on a fixed platform, using  a tripod stand   or
the Thorvald robot \cite{grimstad2017thorvald}, located at the headland of     a
given crop row of the experimental field. 

The image data was bagged at $30\,\mbox{fps}$ rate, where a series of    actions
were recorded in an orderly manner. In \cite{das2018discrete}, the       authors
explain a harvesting protocol for real-time soft-fruit harvesting operation   in
strawberry production fields.
There were five activities mentioned, namely \textit{idle},    \textit{picking},
\textit{loading}, \textit{unloading}, and \textit{transportation}.
We have similarly demonstrated the harvesting operation in an       agricultural
environment, focusing on mainly two activities, i.e., \textit{Picking},      and
\textit{Not Picking}. 
Picking activity deals with actions involving human picker motions,    occurring
while picking up fruits from crops, and
\textit{Not Picking} activity includes all other activities, such             as
\textit{idle}, \textit{unloading}, \textit{loading} and \textit{transportation}. 
%
%
From now on, we call the \textit{Not Picking} activity  as    \textit{unloading}
throughout the manuscript.

%
%
Multiple video data were collected for different agricultural          scenarios
(Fig.\,\ref{fig:data_collection_polyorch}) with varying combinations of     both
activities to promote randomness in the dataset. 
%
%
Then, we have manually processed the raw video data into three types          of
classified video dataset: \textit{picking dataset} with both pickers    picking,
\textit{unloading dataset} with both pickers unloading,                      and
\textit{mixed dataset} with both pickers doing different activities.
In the mixed dataset, several combinations of activities are considered for both
pickers. For example, in a single video clip, one picker might be picking first,
then switch to unloading, while another picker might only be picking         (or
unloading) in the whole video.
%
%
Each classified dataset consists of $6\minus7$ videos, with a frame rate      of
$10\,\,\mbox{fps}$ and a resolution of $640\times480$ pixels, and     $4\minus6$
seconds video chunks (.avi or .MP4) for each activity, wherein every      single
video consists of $40\minus60$ frames in total.

%
\section{Proposed Algorithm Overview}
%
This section will briefly explain each module of our proposed            activity
recognition and classification architecture based on the video dataset  collected
in distinct crop fields.
%
%
%
\vspace{-2.5mm}
\subsection*{Object Detection Module (ODM) using Mask R-CNN}
\begin{figure}[htbp]
\centering
\includegraphics[width=8.4cm]{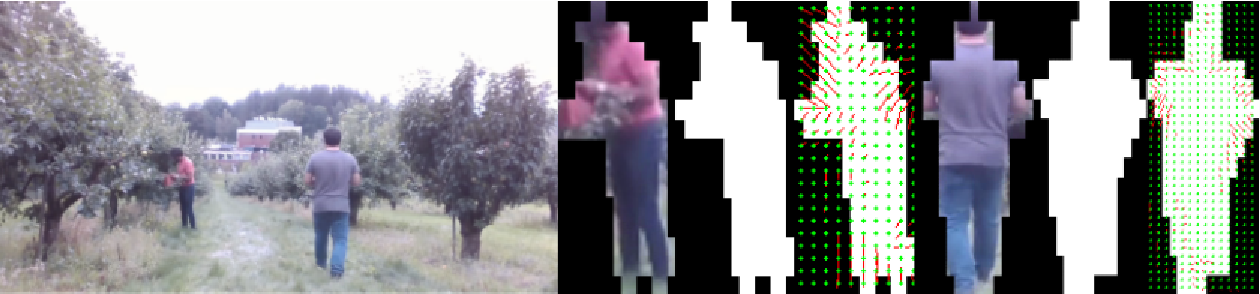}
\caption{From left to right: original frame with two pickers; using Mask   R-CNN
for consecutively detecting pickers, then fetching a MOR in each frame with  its
colored and binary masks; finally, applying optical flow on selected grids.}
%
\label{fig:mask_results}
\end{figure}

This module is used to extract moving targets in frame sequences under    various
visual challenges in datasets such as noise, shadows, poor illumination, and  low
frame resolution.
%
%
%
We emphasize the use of Mask R-CNN architecture \cite{he2017mask} due to      its
ability to robustly handle scene challenges and work precisely at     pixel-level
instance segmentation of moving objects in the images.
%
%
%
%
This network framework provides class labels, bounding boxes, and    segmentation
masks for each detected object.
%
%
We have performed transfer learning on the pre-trained model on COCO      (Common
Objects in Context) dataset.
%
%
%
The CNN model was trained on our dataset using GPU (GTX 1080Ti,   Nvidia,  Corp.)
Desktop with  CUDA 9.0. 
%
%
The trained weights were saved and used in the OpenCV library in Python      for
further application in the classification pipeline. 
Next, each inference input frame fed in the trained network provides    instance
segmentation for foreground class `\textit{person}', with its   predicted box,
overlay colored and binary masks contours inside the boxes                  (see
Fig.\,\ref{fig:mask_results}). 
The extracted foreground mask is called Moving Object Region (MOR), that      is
used by the MEM block.
%
\vspace{-3.5mm}
\subsection*{Motion Estimation Module (MEM) based on Optical flow}
This module estimates motion using an optical flow algorithm on a MOR to   obtain
the flow motion descriptors (i.e., magnitude and orientation) and analyze    them
using statistical attributes.
We use the Gunnar-Farneback optical flow method \cite{farneback2003two}   because
of its ability to look at all of the feature vectors at pixel-level and       its
intensity changes between the two consecutive frames.
%
%
Motion descriptors are feature vectors calculated for each frame, or       several
consecutive frames, and used for analysis.
We have combined the use of Kernel Density Estimation (KDE)              analysis
\cite{ardalan2014} and K-means clustering algorithm to identify the    underlying
critical classes in motion descriptors data.
%
%
%
Then, we have applied the univariate KDE with Gaussian kernel, and the  parameter
bandwidth is determined automatically using the adaptive bandwidth      selection
method based on a linear diffusion process.
%
%

The basis of evaluation in the K-means clustering algorithm is the     silhouette
coefficient method, which has values within the interval            $[\minus1,1]$
\cite{wang2017kmeans}. 
%
%
Then, we want the coefficients to be as high as possible and close to the unit to
have representative clusters. 
%
%
We have analysed motion descriptor data distribution statistically by calculating
their    mean value (\textit{mean}), standard deviation (\textit{std.}),    range
(\textit{range}), minimum (\textit{min.}), maximum (\textit{max.}), and root mean
square (\textit{rms}). 
%
%
Moreover, we have also calculated another feature vector by defining             a
relation between flow magnitude and orientation. Thus, any significant changes  in
motion descriptors can be evidently noticed, which would otherwise not be possible. 
%
%
In this way, we make our algorithm classification criteria   more sensitive to be
able to sparse different patterns effectively. 
This motion descriptor is named in the paper as     Correlation Sensitivity (CS),
and it is defined as: 
%
\begin{equation}\label{eq:7}
{
\mbox{CS} = \sum_{f=1}^F \frac{\mbox{mag}_{f} \cdot \theta_{f}}
    {2 \cdot S(V_f, \theta_{f})} \,,
}
\end{equation}
where, for a set of input MOR frames \textit{F} in a video, $S(V_{f},\theta_{f})$,
$V_{f}$, $mag_{f}$ and $\theta_{f}$ are respectively the total number of     flow
features, the flow magnitude and the flow orientation for each frame $f$. 
We have chosen only those statistical attributes of motion descriptors that  show
a sparse representation of different patterns in KDE analysis. 
%
%
Therefore, among several statistical attributes for each motion descriptor,    we
have relied upon a 4 (four) parameters vector, such as  \textit{range} values for
flow magnitude, (\textit{max.}) and (\textit{min.}) values for flow   orientation
and \textit{mean} value for CS. 
%
%
Finally, we set the threshold value for each parameters vector of the  identified
clusters to provide class labels to different activities. 
%
%
\subsection*{Motion Classification Module (MCM)}
Here, we introduce a hierarchical approach to separate the picker's actions    to
increase the effectiveness of the classification, wherein we are    investigating
the activity classification at two levels: 
(i)
Frame-level (FL) classification, that classifies each frame into  two pre-classes
as \textit{Picking} (P) and \textit{Not Picking} (NP); (ii) Batch-Frame     level
(BFL) classification,  that classifies further batch of FL annotation into    two
classes for showing the continuity of an activity as: if P,                  then
\textit{Wait to Finish}; otherwise, if NP, then \textit{Call a robot}. 
To achieve this, we have developed a classification correction algorithm     that
uses the list of FL class labels in a rolling window spanning the last       five
frames.  
Then, we choose the most occurred FL class label from the sequence of labels   to
create the respective BFL class label. For convenience, we named it as    rolling
mode method.
%
%
The BFL labels are part of the RSM block in the proposed  video-based recognition
and classification framework, indicating future research to solve the       field
logistic problem by scheduling the fleet of robots from picker's         activity
predictions. 
\section{Experimental Results and Discussion}
This section presents the experimental tests and the numerical results   obtained
with the proposed video-based classification framework for picker's      activity
classification on a mixed activity dataset consisting of $600$ frames with    $2$
pickers in each, makes in total $1200$ MOR frames.
%
%
%
%
%
%
%
\subsection*{Activity Classification Analysis}
\begin{figure}[!htbp]
\centering
\includegraphics[width=5.4cm]{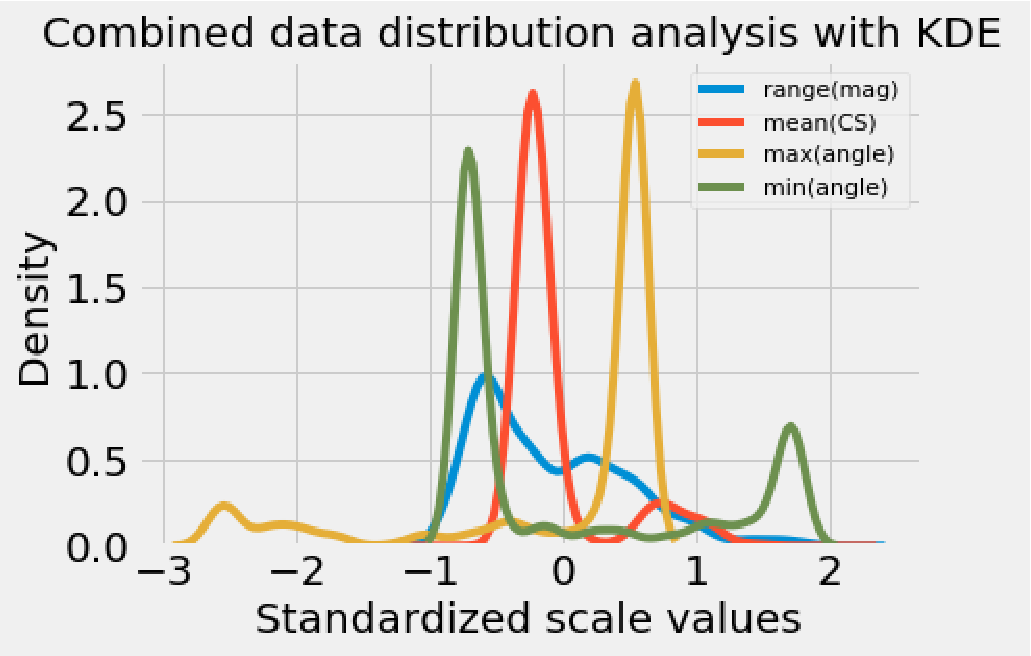}
\caption{KDE distribution analysis of the four parameters vectors,          where
evidently two local maxima are found in each plot considered as two       natural
clusters: (blue) \textit{range} of magnitude; (red) \textit{mean} of CS; (yellow)
\textit{max.} of flow orientation; (green) \textit{min.} of flow orientation.}
%
\label{fig:density_distribution_kde}
\end{figure}

%
%
%
For a given statistical attribute, the combination of the KDE        distribution
analysis and the best evaluation of the K-means silhouette score suggests  mainly
two  clusters ($k\!=\!2$), as shown respectively                               in
Fig.\,\ref{fig:density_distribution_kde} and Table\,\ref{tb:activity_silhouette}. 
The corresponding K-means cluster centers are shown                            in
Table\,\ref{tb:cluster_center}, where NP is                  \textit{Not Picking}
(or \textit{unloading}) and P is \textit{Picking}. 
%
Accordingly, the threshold values are set for all four parameters vectors. 
Next, the FL classification performance is evaluated using statistical   measures
of sensitivity (TPR), specificity (TNR), and accuracy (ACC) of the       activity
detection as the average values over the number of frames. 
%
%
%
%
%
%

\begin{figure}[!htbp]
\centering
    \begin{subfigure}[b]{0.23\textwidth}
    \centering
    \includegraphics[width=1.04\hsize]{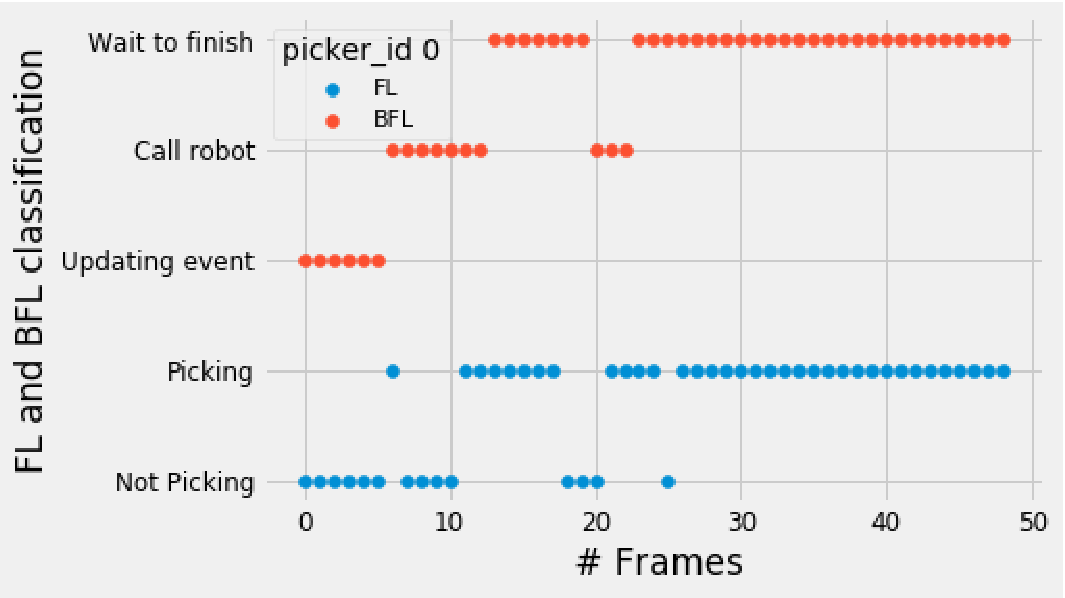}
    \caption{Picker\_00}
    \label{fig:video2_00}
    \end{subfigure} 
    \hspace{-0.2mm}
    \begin{subfigure}[b]{0.23\textwidth}
    \centering
    \includegraphics[width=1.04\hsize]{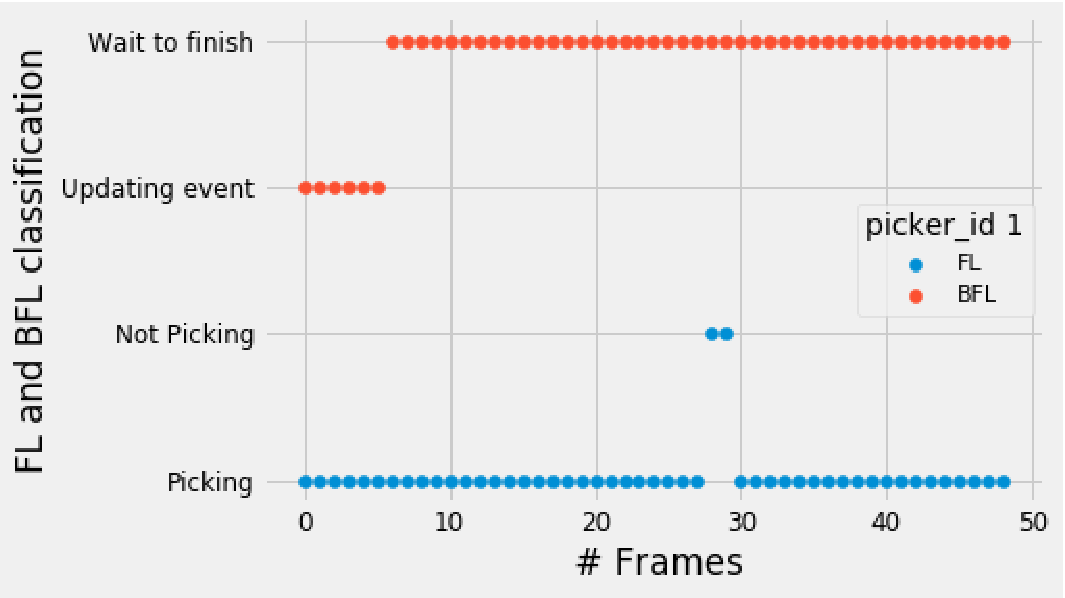}
    \caption{Picker\_01}
    \label{fig:video2_01}
    \end{subfigure} 
    \label{fig:orchard_FL_BLF_results}
    \caption{
    Experimental results of activity classification with FL and  BLF approaches from
    test dataset.} 
\end{figure}

\begin{table}[!htbp]
\caption{Activity classification: K-means silhouette coefficients for different statistical attributes and size of clusters.}
\centering
\begin{tabular}{||c|c|c|c|c||}
\hline
\textbf{\textit{k}} & \textbf{\textit{range}} & \textbf{\textit{mean}} & \textbf{\textit{max.}} & \textbf{\textit{min.}} \\
&\textbf{(flow mag.)} & \textbf{(CS)} & \textbf{(flow ori.)} & \textbf{(flow ori.)} \\  \hline
2 & 0.65 & 0.90 & 0.87 & 0.87  \\
3 & 0.60 & 0.87 & 0.87 & 0.85  \\
4 & 0.57 & 0.60 & 0.85 & 0.85  \\ \hline
\end{tabular}
\label{tb:activity_silhouette}
\end{table}
\begin{table}[!htbp]
\caption{Activity classification: K-means cluster center.}
\centering
\begin{tabular}{||c|c|c|c|c||} \hline
\textbf{\textit{k}} & \textbf{\textit{range}} &  \textbf{\textit{mean}} & \textbf{\textit{max.}} & \textbf{\textit{min.}}  \\
                    & \textbf{\textbf{(flow mag.)}} &  \textbf{(CS)}          & \textbf{(flow ori.)}   & \textbf{(flow ori.)}    \\
\hline
NP & 1550.49 & 3433.18 & 176.55 &  3.47 \\
P &  708.78 &  675.95 & 104.82 & 79.65 \\
\hline
units & \textit{pixel}        & \textit{pixel}     & \textit{degree}            & \textit{degree} \\ \hline 
\end{tabular}
\label{tb:cluster_center}
\end{table}
\begin{table}[!htbp]
\caption{Activity classification using the proposed classification      algorithm
and its different variants.}
\centering
\begin{tabular}{||c|c|c|c||} \hline
\textbf{classification methods} & \textbf{ACC}\,\% & \textbf{TNR}\,\% & \textbf{TPR}\%  \\ \hline
proposed solution  	   & 84.18 & 89.06 & 75.00 \\
variant (i)            & 50.51 & 90.23 & 40.12 \\
variant (ii)           & 66.33 & 94.52 & 49.59 \\
variant (iii)          & 60.20 & 82.72 & 44.35 \\ \hline
\end{tabular}
\label{tb:Mixed_activity}
\end{table}
\vspace{-5mm}
We have also analysed our classification framework with various combinations   of
all four parameters vectors named as variants and listed as: variant (i),   which
includes \textit{range} value of the flow magnitude, as well as \textit{min.} and
\textit{max.} values of the flow orientation, that is, (\textit{range})      flow
magnitude + (\textit{min.},\,\textit{max.}) flow orientation; variant (ii), which
includes (\textit{range}) value of the flow magnitude; and variant (iii),   which
includes (\textit{min.},\,\textit{max.}) values of the flow orientation.
%
%
%
%
%
The proposed solution presents the higher ACC and TPR values respectively      as
$84\,\%$ and  $75$\,\%, as shown in Table\,\ref{tb:Mixed_activity}, revealing the
added advantage of including the CS feature vector for activity    classification
compared to other algorithm variants without using CS.
%
%
%
%
An example of the performance analysis of our proposed FL and BFL  classification
algorithm on test videos is shown in Figures\,\ref{fig:video2_00}             and
\ref{fig:video2_01}, which depict the human pickers as $picker\_ID\,00$       and
$picker\_ID\,01$ respectively.
%
%
%
The BFL classification for the corresponding FL classification shows an   overall
improvement in classifying the activity duration. Meanwhile, we argue that    the
MOR mask has successfully detected the foreground moving object amidst    various
scene challenges.
\section{Concluding Remarks}
In this work, a video-based activity classification framework is proposed    that
uses the combination of Mask R-CNN architecture and optical flow           motion
descriptors. 
%
%
We have defined four modules in this framework with their distinct functions   in
classification pipeline. 
%
%
%
The proposed classification criteria with a newly derived Correlation Sensitivity
(CS) shows satisfactory classification results at a Frame Level (FL)         with
accuracy, specificity and sensitivity as close as $84\,\%$, $89\,\%$ and $75\,\%$,
respectively.
A running-mode algorithm also shows a good classification correction at a   Batch
Frame Level (BFL). 
%
%
%
The developed video-based benchmark model shows satisfactory overall results under
variability in shading, brightness, blur of images, and occlusion.
The accuracy analysis of the classification framework in a multiple        pickers
scenario using Deep Neural Networks (DNNs) can be explored in future research.
%
%
%
%
\bibliographystyle{IEEEtran}
\bibliography{references}  
%
\end{document}